\documentclass[10pt,twocolumn,letterpaper]{article}

\usepackage{iccv}
\usepackage{times}
\usepackage{epsfig}
\usepackage{graphicx}
\usepackage{amsmath}
\usepackage{amssymb}
\usepackage{caption}
\usepackage{color}
\usepackage{soul}
\usepackage{bbm}
\usepackage{booktabs}

\newcommand{\system}{FiNet\xspace}
\iccvfinalcopy %
\def\iccvPaperID{10086} %

\definecolor{citecolor}{RGB}{34,139,34}
\newcommand{\ra}[1]{\renewcommand{\arraystretch}{#1}}

\usepackage[pagebackref=true,breaklinks=true,letterpaper=true,colorlinks,citecolor=citecolor,bookmarks=false]{hyperref}

\begin{document}

\title{Compatible and Diverse Fashion Image Inpainting}

\author{Xintong Han$^{1,2}$ \quad Zuxuan Wu$^3$ \quad Weilin Huang$^{1,2}$ \quad Matthew R. Scott$^{1,2}$ \quad Larry S. Davis$^3$\\
$^1$Malong Technologies, Shenzhen, China \\
$^2$Shenzhen Malong Artificial Intelligence Research Center, Shenzhen, China \\
$^3$University of Maryland, College Park \\
}

\twocolumn[{%
\renewcommand\twocolumn[1][]{#1}%
\maketitle

\begin{center}
\vspace{-20pt}
\centering
\includegraphics[width=0.98\textwidth]{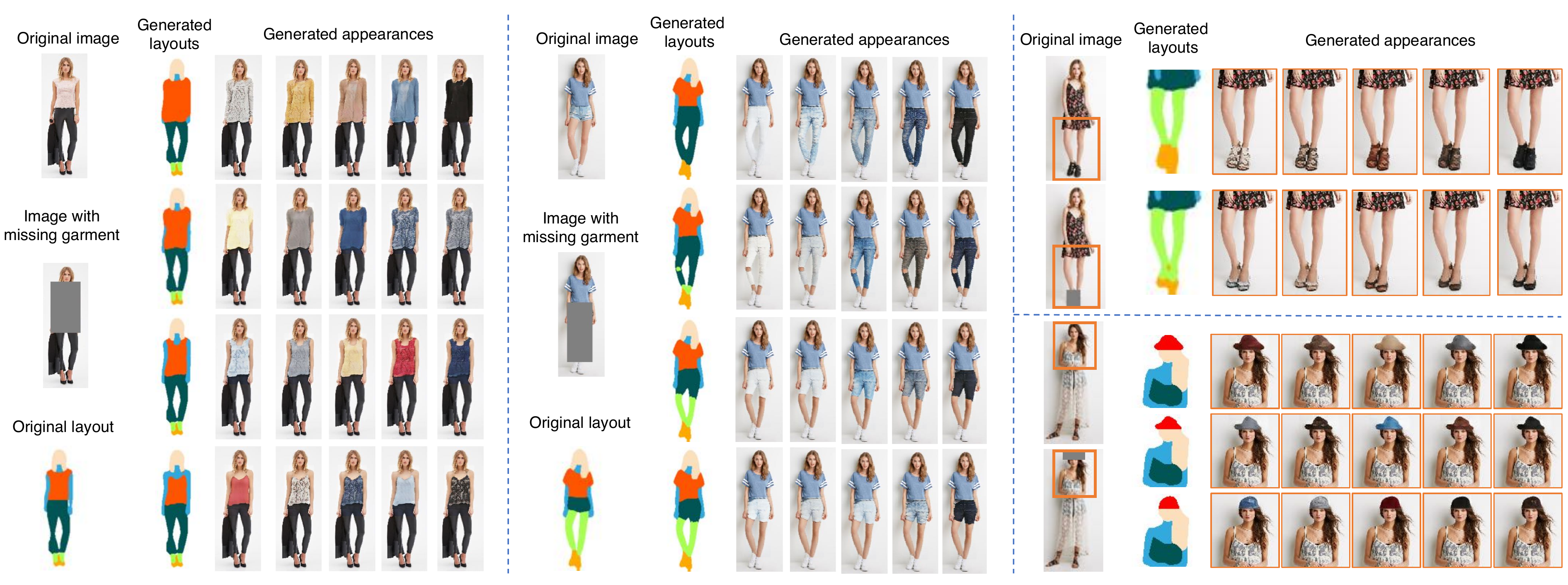}
\vspace{-5pt}
\captionof{figure}{We inpaint missing fashion items with compatibility and diversity in both shapes and appearances.}
\label{fig:teaser}
\end{center}
}]
\vspace{6pt}

\begin{abstract}
Visual compatibility is critical for fashion analysis, yet is missing in existing fashion image synthesis systems. In this paper, we propose to  
explicitly model visual compatibility through fashion image inpainting. To this end, we present Fashion Inpainting Networks (FiNet), a two-stage image-to-image generation framework that is able to perform compatible and diverse inpainting. Disentangling the generation of shape and appearance to ensure photorealistic results, our framework consists of a shape generation network and an appearance generation network. More importantly, for each generation network, we introduce two encoders interacting with one another to learn latent code in a shared compatibility space. The latent representations are jointly optimized with the corresponding generation network to condition the synthesis process, encouraging a diverse set of generated results that are visually compatible with existing fashion garments. In addition, our framework is readily extended to clothing reconstruction and fashion transfer, with impressive results. Extensive experiments on fashion synthesis task quantitatively and qualitatively demonstrate the effectiveness of our method.

\end{abstract}

\section{Introduction}
Recent breakthroughs in deep generative models, especially Variational Autoencoders (VAEs) \cite{kingma2013auto}, Generative Adversarial Networks (GANs) \cite{goodfellow2014generative}, and their variants \cite{pix2pix2016,mirza2014conditional,esser2018variational,larsen2016autoencoding}, open a new door to a myriad of fashion applications in computer vision, including fashion design \cite{kang2017visually,sbai2018design}, language-guided fashion synthesis \cite{zhu2017be,rostamzadeh2018fashion,gunel2018language}, virtual try-on systems \cite{han2018viton,wang2018toward,chou2018PIVTONS}, clothing-based appearance transfer \cite{raj2018swapnet,zanfir2018human}, \etc. Unlike generating images of rigid objects, fashion synthesis is more complicated as it involves multiple clothing items that form a compatible outfit. Items in the same outfit might have drastically different appearances like texture and color (\eg, cotton shirts, denim pants, leather shoes, \etc), yet they are complementary to one another when assembled together, constituting a stylish ensemble for a person. Therefore, exploring compatibility among different garments, an integral collection rather than isolated elements, to synthesize a diverse set of fashion images is critical for producing satisfying virtual try-on experiences and stunning fashion design portfolios. However, modeling visual compatibility in computer vision tasks is difficult as there is no ground-truth annotation whether fashion items are compatible. Hence, researchers mitigate this issue by leveraging \textit{contextual relationships} (or co-occurrence) as a form of weak compatibility signal \cite{han2017learning,veit2015learning,song2018neural}. For example, two fashion items in the same outfit are considered as compatible, while those are not usually worn together are incompatible. 

In this spirit, we consider explicitly exploring visual compatibility relationships as contextual clues for the task of fashion image synthesis. In particular, we formulate this problem as image inpainting, which aims to fill in a missing region in an image based on its surrounding pixels.  Note that generating an entire outfit while modeling visual compatibility among different garments at the same time is extremely challenging, as it requires to render clothing items varying in both shape and appearance onto a person. Instead, we take the first step to model visual compatibility by narrowing it down to image inpainting, using images with people in clothing. The goal is to render a diverse set of realistic clothing items to fill in the region of a missing item in an image, while matching the style of existing garments {\color {black} as shown in Figure \ref{fig:teaser}}. This can be readily used for various fashion applications like fashion recommendations, fashion design, and garment transfer. {\color{black}For example, the inpainted item can serve as an intermediate result (\eg, query on Google/Pinterest, picture shown to fashion stylers) to retrieve similar items from the database for recommendation}.

Unlike inpainting a missing region surrounded by rigid objects~\cite{pathak2016context,yeh2017semantic,yu2018generative}, synthesizing a clothing item that is matched with its surrounding garments is more challenging since (1) we wish to generate a diverse set of results, yet the diversity is constrained by visual compatibility; (2) more importantly, the generalization process is essentially a multi-modal problem---given a fashion image with one missing garment, various items, different in both shape and appearance, can be generated to be compatible with the existing set. For instance, in the second example in Figure \ref{fig:teaser}, one can have different types of bottoms in shape (\eg, shorts or pants), and each bottom type may have various colors in visual appearance (\eg, blue, gray or black). Thus, the synthesis of a missing fashion item requires modeling of both shape and appearance. However, coupling their generation simultaneously usually fails to handle clothing shapes and boundaries, thus creating unsatisfactory results \cite{Lassner_2017_ICCV,zhu2017be}.

\begin{figure}
\begin{center}
   \includegraphics[width = \linewidth]{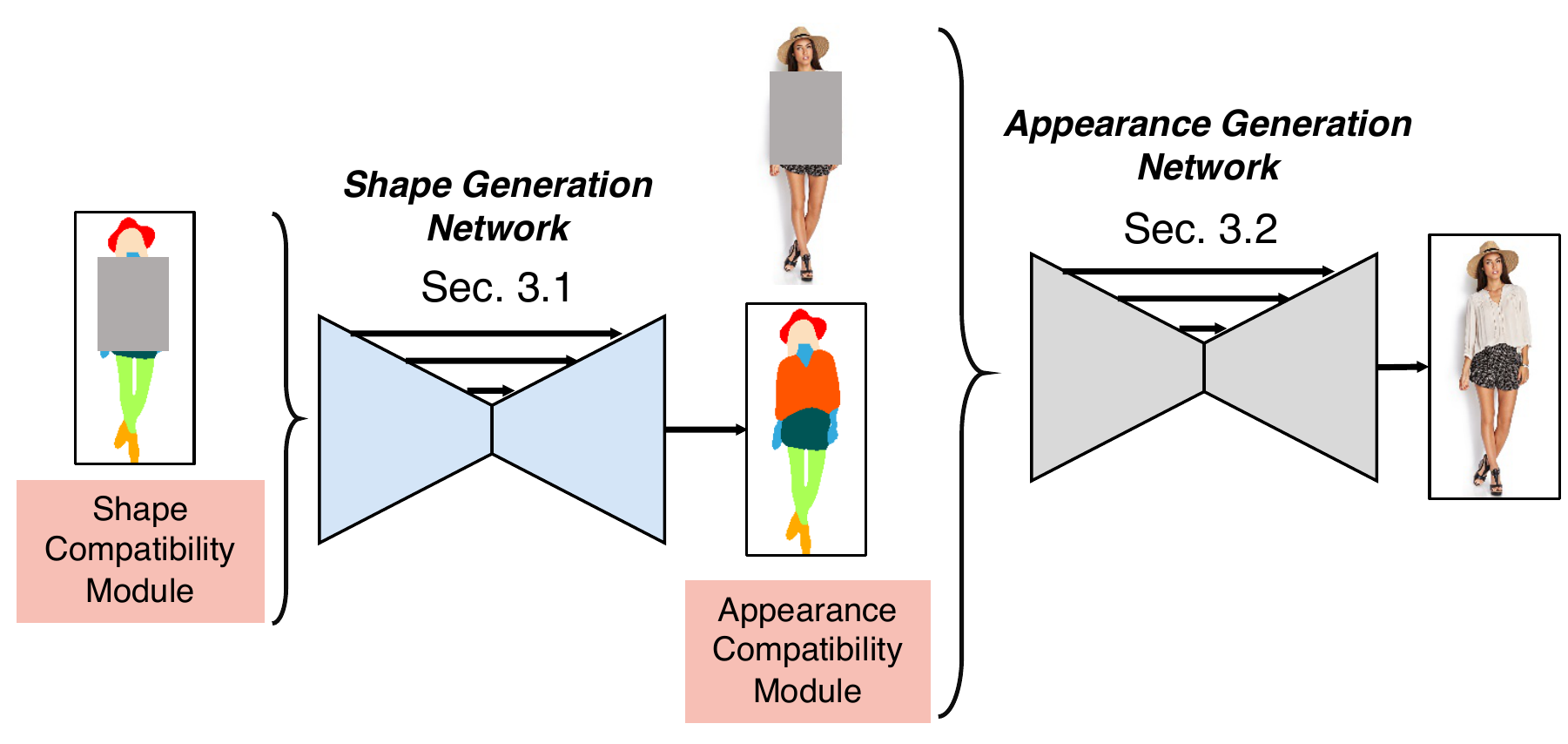}
\end{center}
\vspace{-18pt}
\caption{FiNet framework. The shape generation network (Sec. \ref{sec:31}) aims to fill a missing segmentation map given shape compatibility information, and the appearance generation network (Sec. \ref{sec:32}) uses the inpainted segmentation map and appearance compatibility information for generating the missing clothing regions. Both shape and appearance compatibility modules carry uncertainty, allowing our network to generate diverse and compatible fashion items.}

\label{fig:framework}
\end{figure}

To address these issues, we propose \system,  a two-stage framework illustrated in Figure~\ref{fig:framework}, which fills in a missing fashion item in an image at the pixel-level through generating a set of realistic and compatible fashion items with diversity. In particular, we utilize a shape generation network and an appearance generation network to generate shape and appearance sequentially. Each generation network contains a generator that synthesizes new images through reconstruction, and two encoder networks interacting with each other to encourage diversity while preserving visual compatibility. With one encoder learning a latent representation of the missing item, the second encoder regularizes the latent representation with the latent codes from the second encoder, whose inputs are from neighboring garments ({\color{black} compatible context}) of the missing item. These latent representations are jointly learned with the corresponding generator to condition the generation process.
This allows both generation networks to learn high-level compatibility correlations among different garments, enabling our framework to produce synthesized fashion items with meaningful diversity (multi-modal outputs) and strong compatibility, as shown in Figure \ref{fig:teaser}. We provide extensive experimental results on the DeepFashion \cite{liu2016deepfashion} dataset, with comparisons to state-of-the-art approaches on fashion synthesis, and the results confirm the effectiveness of our method.

\section{Related Work}
\noindent \textbf{Visual Compatibility Modeling.}
Visual compatibility plays an essential role in fashion recommendation and retrieval \cite{liu2012hi,simo2015neuroaesthetics,song2018neural,song2017neurostylist}. %
Metric learning based methods have been adopted to solve this problem by projecting two compatible fashion items close to each other in a style space \cite{mcauley2015image,veit2015learning,vasileva2018learning}. Recently, beyond modeling pairwise compatibility, sequence models \cite{han2017learning,li2016mining} and subset selection algorithms \cite{hsiao2018creating} capable of capturing the compatibility among a collection of garments have also been introduced. Unlike these approaches which attempt to estimate fashion compatibility, we incorporate compatibility information into an image inpainting framework that generates a fashion image containing complementary garments.
Furthermore, most existing systems rely heavily on manual labels for supervised learning. In contrast, we train our networks in a self-supervised manner, assuming that multiple fashion items in an outfit presented in the original catalog image are compatible to each other, since such catalogs are usually designed carefully by fashion experts. Thus, minimizing a reconstruction loss jointly during learning to inpaint can learn to generate compatible fashion items.

\noindent\textbf{Image Synthesis.}
There has been a growing interest in image synthesis with GANs \cite{goodfellow2014generative} and VAEs \cite{kingma2013auto}. To control the quality of generated images with desired properties, various supervised knowledge or conditions like class labels \cite{odena2016conditional,brock2018large}, attributes \cite{shen2016learning,yan2016attribute2image},
text \cite{reed2016generative,zhang2016stackgan,xu2017attngan}, images \cite{pix2pix2016,wang2017high}, \etc, are used. In the context of generating fashion images, existing fashion synthesis methods often focus on rendering clothing conditioned on poses \cite{ma2017pose,neverova2018deep,Lassner_2017_ICCV,siarohin2018deformable}, textual descriptions \cite{zhu2017be,rostamzadeh2018fashion}, textures \cite{xian2018texturegan}, a clothing product image \cite{han2018viton,wang2018toward,yoo2016pixel,jetchev2017conditional}, clothing on another person \cite{zanfir2018human,raj2018swapnet}, or multiple disentangled conditions \cite{esser2018variational,ma2018disentangled,yildirim2018disentangling}. In contrast, we make our generative model aware of fashion compatibility, which has not been explored previously. To make our method more applicable to real-world applications, we formulate the modeling of fashion compatibility as a compatible inpainting problem that captures high-level dependencies among various fashion items or fashion concepts.

Furthermore, fashion compatibility is a many-to-many mapping problem, since one fashion item can match with multiple related items of various shapes and appearances. Therefore, our method is related to multi-modal generative models \cite{zhu2017toward,lee2018diverse,ghosh2017multi,huang2018multimodal,wang2017high,esser2018variational}. In this work, we propose to learn a compatibility latent space, where the compatible fashion items are encouraged to have similar distributions.

\noindent\textbf{Image Inpainting.}
Our method is also closely related to image inpainting \cite{pathak2016context,yeh2017semantic,iizuka2017globally,yu2018generative}, which synthesizes missing regions in an image given contextual information. Compared with traditional image inpainting, our task is more challenging---we need to synthesize realistic fashion items with meaningful diversity in shape and appearance, and at the same time ensure that the inpainted clothing items are compatible in fashion style to existing garments in the current image. This requires explicitly encoding the compatibility by learning inherent relationships between various garments, rather than simply modeling the context itself. Another significant difference is that people expect multi-modal outputs in fashion image synthesis, whereas traditional image inpainting is typically a uni-modal problem.

\begin{figure}
\begin{center}
   \includegraphics[width = 1.0\linewidth]{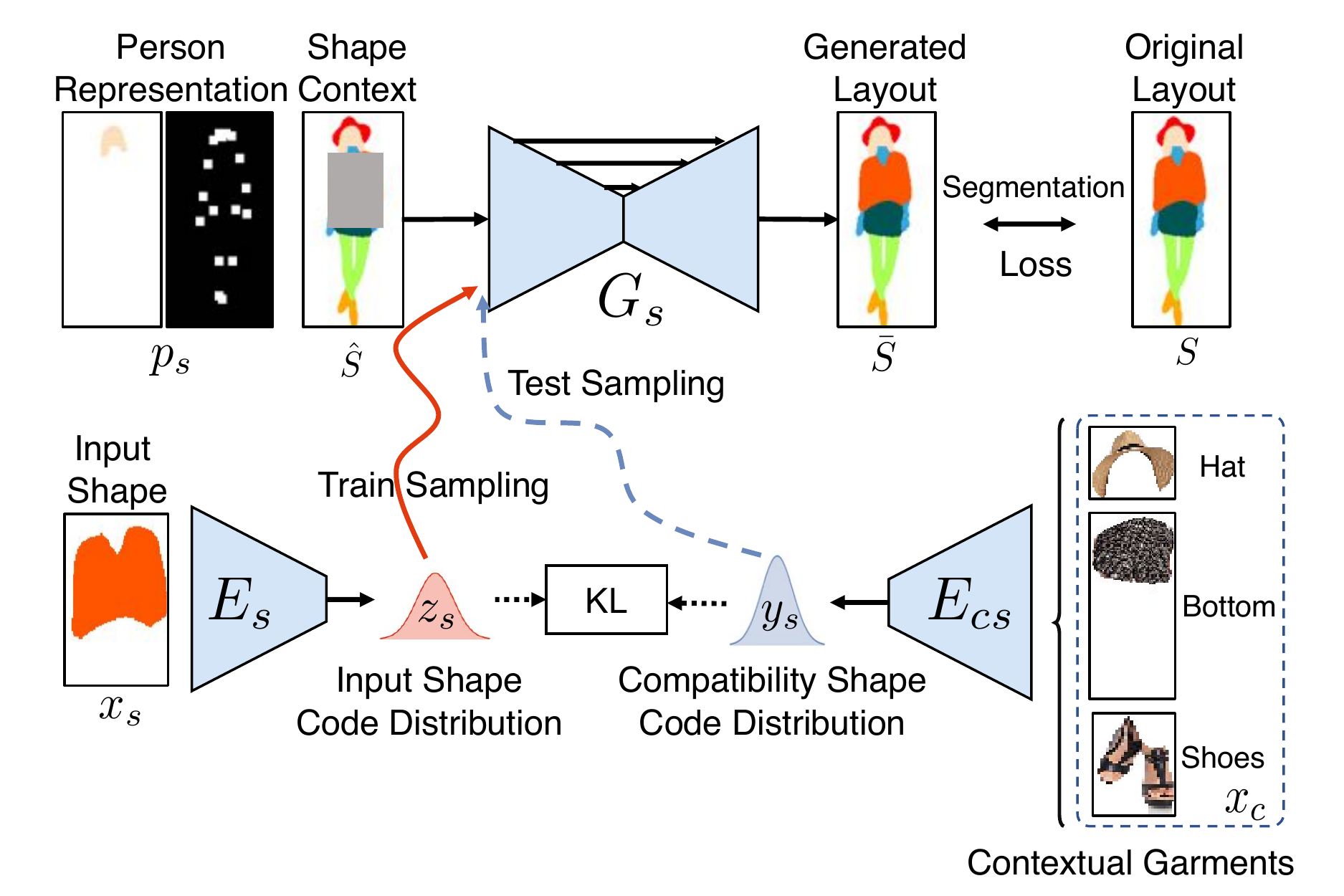}
\end{center}
\vspace{-20pt}
\caption{Our shape generation network.}
\label{fig:shape_generator}
\end{figure}

\section{FiNet: Fashion Inpainting Networks}
Our task is, given an image with a missing fashion item (\eg, by deleting the pixels in the corresponding area), we explicitly explore visual compatibility among neighboring fashion garments to fill in the region, synthesizing a diverse set of photorealistic clothing items varying in both shape (\eg, maxi, midi, mini dresses) and appearance (\eg, solid color, floral, dotted, \etc). Each synthesized result is expected not only to blend seamlessly with the existing image but also to be compatible with the style of current garments (see Figure \ref{fig:teaser}). As a result, these generated images can be readily used for tasks like fashion recommendation. Further, in contrast to rigid objects, clothing items are usually subject to severe deformations, making it difficult to simultaneously synthesize both shape and appearance without introducing unwanted artifacts. To this end, we propose a two-stage framework named Fashion Inpainting Networks (FiNet) that contains a shape generation network (Sec \ref{sec:31}) and an appearance generation network (Sec \ref{sec:32}), to encourage diversity while preserving visual compatibility when filling in missing region.
Figure \ref{fig:framework} illustrates an overview of the proposed framework. In the following, we present the components of FiNet in detail.

\subsection{Shape Generation Network}
\label{sec:31}
Figure~\ref{fig:shape_generator} shows an overview of the shape generation netowrk. It contains an encoder-decoder based generator $G_s$ to synthesize a new image through reconstruction, and two encoders, working collaboratively to condition the generation process, producing compatible synthesized results with diversity. More formally, the goal of the shape generation network is to learn a mapping with $G_s$ that projects a shape context with a missing region $\hat{S}$ as well as a person representation $p_s$ to  a complete shape map $S$, conditioned on the shape information captured by a shape encoder $E_s$. 

To obtain the shape maps for training the generator, we leverage an off-the-shelf human parser \cite{gong2018instance} pretrained on the Look Into Person dataset \cite{gong2017look}. In particular, given an input image $I \in R^{H \times W \times 3}$, we first obtain its segmentation maps with the parser, and then re-organize the parsing results into 8 categories:  \textit{face and hair}, \textit{upper body skin} (torso + arms), \textit{lower body skin} (legs), \textit{hat}, \textit{top clothes} (upper-clothes + coat), \textit{bottom clothes} (pants + skirt + dress), \textit{shoes} \footnote{We only consider 4 types of garments: hat, top, bottom, shoes in this paper, but our method is generic and can be extended to more fine-grained categories if segmentation masks are accurately provided.}, and \textit{background} (others). The 8-category parsing results are then transformed into an 8-channel binary map $S \in \{0,1\}^{H \times W \times 8}$, which is used as the ground truth of the reconstructed segmentation maps for the input. The input shape map $\hat{S}$ with a missing region is generated by masking out the area of a specific fashion item in the ground truth maps.  For example, in Figure \ref{fig:shape_generator}, when synthesizing top clothes, the shape context $\hat{S}$ is produced by masking out the plausible top region, represented by a bounding box covering the regions of the top and upper body skin. 

In addition, to preserve the pose and identity information in shape reconstruction, we employ similar clothing-agnostic features $p_{s}$ as described in \cite{han2018viton,wang2018toward}, which includes a pose representation, and the hair and face layout. More specifically, the pose representation contains an 18-channel heatmap extracted by an off-the-shelf pose estimator \cite{chen2017cascaded} trained on the COCO keypoints detection dataset \cite{lin2014microsoft}, and the face and hair layout is computed from the same human parser~\cite{gong2018instance} represented by a binary mask whose pixels in the face and hair regions are set to 1. Both representations are then concatenated to form $p_{s} \in \rm R^{H \times W \times C_s}$, where $C_s=18+1=19$ is the number of channels.

Directly using $\hat S$ and $p_s$ to reconstruct $S$, \ie, $G_s(\hat S, p_s)$, using standard image-to-image translation networks \cite{pix2pix2016,ma2017pose,han2018viton}, although feasible, will lead to a unique output without diversity. We draw inspiration from variational autoencoders, and further condition the generation process with a latent vector $z_s \in \rm R^\mathrm{Z}$, that encourages diversity through sampling during inference. As our goal is to produce various shapes of clothing items to fill in a missing region, we train $z_s$ to encode shape information with $E_s$. Given an input shape $x_s$ ($x_s$ is the ground truth binary segmentation map of the missing fashion item obtained by $S$), the shape encoder $E_s$ outputs $z_s$, leveraging a re-parameterization trick to enable a differentiable loss function~\cite{zhu2017toward,dosovitskiy2016generating}, \ie, $z_s \sim E_s(x_s)$. $z_s$ is usually forced to follow a Gaussian distribution $\mathcal{N}(0, \mathbbm{1})$ during training, which enables stochastic sampling at the test time when $x_s$ is unknown:
\begin{equation}
L_{KL} = D_{KL}(E_s(x_s)~||~\mathcal{N}(0,\mathbbm{1})),
\label{eqn:kl}
\end{equation}
where $D_{KL}(p||q) = \int p(z)\log\frac{p(z)}{q(z)}dz$ is the KL divergence. The learned latent code $z_s$, together with the shape context $\hat{S}$ and person representation $p_s$ are input to the generator $G_s$ to produce a complete shape map with missing regions filled: $\bar S = G_{s}(\hat S, p_{s}, z_s)$. Further, the shape generator is optimized by minimizing the cross entropy segmentation loss between $\bar S$ and $S$:
\begin{equation}
L_{seg} = -\frac{1}{H W}\sum_{m=1}^{H W}\sum_{c=1}^{C} S_{mc}\log(\bar S_{mc}),
\end{equation}
where $C=8$ is the number of channels of the segmentation map. The shape encoder $E_s$ and the generator $G_s$ can be optimized jointly by minimizing:
\begin{equation}
L = L_{seg} + \lambda_{KL}L_{KL},
\label{eq:cvae}
\end{equation}
where $\lambda_{KL}$ is a weight balancing two loss terms. At test time, one can directly sample from $\mathcal{N}(0,\mathbbm{1})$ to generate $z_s$, enabling the reconstruction of a diverse set of results with $\bar S = G_{s}( \hat S, p_{s}, z_s)$.

Although the shape generator now is able to synthesize different garment shapes, it fails to consider visual compatibility relationships. Consequently, many generated results are visually unappealing (as will be shown in experiments). 
To mitigate this problem, we constrain the sampling process via modeling the visual compatibility relationships using  
existing fashion garments in the current image, which we refer to as \textit{contextual garments}, denoted as $x_c$. To this end, we introduce a shape compatibility encoder $E_{cs}$, with the goal of learning the correlations between the shapes of synthesized garments and contextual garments. 

{\color {black} This intuition is based on the same assumption in compatibility modeling approaches that fashion items usually worn together are compatible \cite{han2017learning,veit2015learning,song2018neural}, and hence the contextual garments (co-occurred garments) contain rich compatibility information about the missing item. As a result, if } a fashion garment is compatible with those contextual garments, its shape can be determined by looking at the context. For instance, given a men's tank top in the contextual garments, the synthesized shape of the missing garment is more likely to be a pair of men's shorts than a skirt. The idea is conceptually similar to two popular models in the text domain, \ie, continuous bag-of-words (CBOW)~\cite{mikolov2013distributed} and skip-gram models~\cite{mikolov2013distributed}; learning to predict the representation of a word given the representations of contextual words around it and vice versa.

{\color {black} As shown in Figure \ref{fig:shape_generator}}, we first extract image segments of contextual garments using $S$. Then, we form the visual representations of the contextual garments $x_c$ by concatenating these image segments from hat to shoes. 
The compatibility encoder $G_{cs}$ then projects $x_c$ into a compatibility latent vector $y_s$, \ie, $y_s \sim E_{cs}(x_c)$.
In order to use $y_s$ as a prior for generating $\bar S$, we posit that a target garment $x_s$ and its contextual garments $x_c$ should share the same latent space. This is similar to the shared latent space assumption applied in unpaired image-to-image translation \cite{liu2017unsupervised,huang2018multimodal,lee2018diverse}). Thus, the KL divergence in Eqn. \ref{eqn:kl} can be modified as,
\begin{equation}
\hat L_{KL} = D_{KL}(E_s(x_s)~||~E_{cs}(x_c)),
\label{eqn:kl2}
\end{equation}
which penalizes the distribution of $z_s$ encoded by $E_s(x_s)$ for being too far from its compatibility latent vector $y_s$ encoded by $E_{cs}(x_c)$.	The shared latent space of $z_s$ and $y_s$ can be also considered as a \textit{compatibility space}, which is similar in spirit to modeling pairwise compatibility using metric learning \cite{veit2015learning,vasileva2018learning}. Instead of reducing the distance between two compatible samples, we minimize the difference between two distributions as we need randomness for generating diverse multi-modal results. 
Through optimizing Eqn. \ref{eqn:kl2}, the generation of $\bar S = G_{s}(\hat S, p_s, z_s)$ not only is aware of the inherent compatibility information embedded in contextual garments, but also enables compatibility-aware sampling during inference when $x_s$ is not available---we can simply sample $y_s$ from $E_{cs}(x_c)$, and compute the final synthesized shape map using $\bar S = G_{s}(\hat S, p_s, y_s)$. Consequently, the generated clothing layouts should be visually compatible to existing contextual garments.
The final objective function of our shape generation network is 
\begin{equation}
L_{s} = L_{seg} + \lambda_{KL}\hat L_{KL}.
\end{equation}

\subsection{Appearance Generation Network}
\label{sec:32}
\begin{figure}
\begin{center}
   \includegraphics[width = 1.0\linewidth]{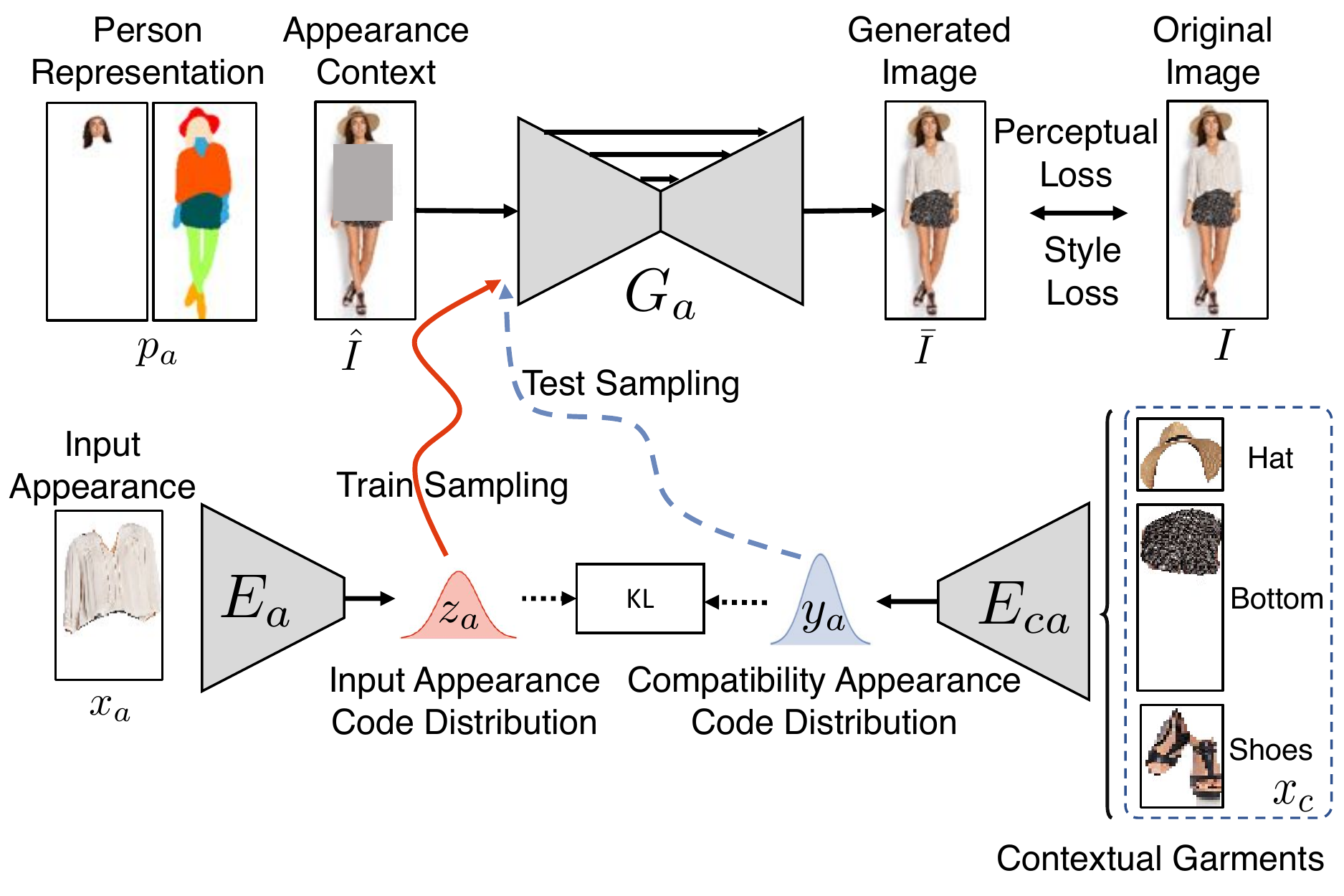}
\end{center}
\vspace{-20pt}
\caption{Our appearance generation network.}
\label{fig:appearance_generator}
\end{figure}

As illustrated in Figure \ref{fig:framework}, the generated compatible shapes of the missing item are input into the appearance generation network to generate compatible appearances. The network has an almost identical structure as our shape generation network, consisting of an encoder-decoder generator $G_{a}$ for reconstruction, an appearance encoder $E_{a}$ that encodes the desired appearance into a latent vector $z_a$, and an appearance compatibility encoder $E_{ca}$ that projects the appearances of contextual garments into a latent appearance compatibility vector $y_a$. Nevertheless, the appearance generation differs from the one modeling shapes in the following aspects. First, as shown in Figure \ref{fig:appearance_generator}, the appearance encoder $E_a$ takes the appearance of a missing clothing item as input instead of its shape, to produce a latent appearance code $z_a$ as input to $G_a$ for appearance reconstruction.

In addition, unlike $G_s$ that reconstructs a segmentation map by minimizing a cross entropy loss, the appearance generator $G_a$ focuses on reconstructing the original image $I$ in RGB space, given the appearance context $\hat I$, in which the fashion item of interest is missing. Further, the person representation $p_{a} \in  \rm R ^ {H\times W\times 11}$ that is input to $G_{a}$ consists of the ground truth segmentation map $S \in \rm R ^ {H\times W\times 8}$ (at test time, we use the segmentation map $\bar S$ generated by our first stage as $S$ is not available), as well as a face and hair RGB segment. The segmentation map contains richer information than merely using keypoints about the person's configuration and body shape, and the face and hair image constrains the network to preserve the person's identity in the reconstructed image $\bar I = G_{a}(\hat I, p_{a}, z_a)$. To reconstruct $I$ from $\bar I$, we adopt the losses widely used in style transfer~\cite{johnson2016perceptual,wu2018dcan,xian2018texturegan}, which contains a perceptual loss that minimizes the distance between the corresponding feature maps of $I$ and $\bar I$ in a perceptual neural network, and a style loss that matches their style information:
\begin{equation}
L_{rec} = \sum_{l=0}^5\lambda_l||\phi_l(I) - \phi_l(\bar I)||_1 +
\sum_{l=1}^5\gamma_l||\mathcal G_l(I) - \mathcal G_l(\bar I)||_1,
\label{eq:loss_app}
\end{equation}
where $\phi_l(I)$ is the $l$-th feature map of image $I$ in a VGG-19 \cite{simonyan2014very} network pre-trained on ImageNet. When $l \geq 1$, we use \texttt{conv1\_2}, \texttt{conv2\_2}, \texttt{conv3\_2}, \texttt{conv4\_2}, and \texttt{conv5\_2} layers in the network, while $\phi_0(I)=I$.
In the second term, $\mathcal G_l \in \mathbb{R}^{C_l\times C_l}$ is the Gram matrix \cite{gatys2016image}, which calculates the inner product between vectorized feature maps:

\begin{equation}
\mathcal G_l(I)_{ij}  = \sum_{k=1}^{H_l W_l}\phi_l(I)_{ik}\phi_l(I)_{jk}~, %
\end{equation}
where $~\phi_l(I) \in \mathbb{R}^{C_l \times H_l W_l}$ is the same as in the perceptual loss term, and $C_l$ is its channel dimension. $\lambda_l$ and $\gamma_l$ in Eqn. \ref{eq:loss_app} are hyper-parameters balancing the contributions of different layers, and are set automatically following \cite{han2018viton,chen2017photographic}. By minimizing Eqn. \ref{eq:loss_app}, we encourage the reconstructed image to have similar high-level contents as well as detailed textures and patterns as the original image.

In addition, to encourage diversity in synthesized appearance (\ie, different textures, colors, \etc), we leverage an appearance compatibility encoder $E_{ca}$, taking the contextual garments $x_c$ as inputs to condition the generation by a KL divergence term $\hat L_{KL} = D_{KL}(E_a(x_a)~||~E_{ca}(x_c))$. The objective function of our appearance generation network is:
\begin{equation}
L_{a} = L_{rec} + \lambda_{KL}\hat L_{KL}.
\end{equation}
Similar to the shape generation network, our appearance generation network, by modeling appearance compatibility, can render a diverse set of visually compatible appearances conditioned on the generated clothing segmentation map and the latent appearance code during inference: $\bar I = G_a(\hat I, p_a, y_a)$, where $y_a \sim E_{ca}(x_c)$.

\subsection{Discussion}
\label{sec:diss}
While sharing the exact same network architecture and inputs, the shape compatibility encoder $E_{cs}$ and the appearance compatibility encoder  $E_{ca}$ model different aspects of compatibility; therefore, their weights are not shared. During training, we use ground truth segmentation maps as inputs to the appearance generator to reconstruct the original image. During inference, we first generate a set of diverse segmentations using the shape generation network. Then, conditioned on these generated semantic layouts, the appearance generation network renders textures onto them, resulting in compatible synthesized fashion images with both diverse shapes and diverse appearances. Some examples are presented in Figure \ref{fig:teaser}. In addition to compatibly inpainting missing regions with meaningful diversity trained with reconstruction losses, our framework also has the ability to render garments onto people with different poses and shapes as will be demonstrated in Sec \ref{sec:rec}.

Note that our framework does not involve adversarial training \cite{pix2pix2016,liu2017unsupervised,ma2017pose,zhu2017be} (hard to stabilize the training process) or bidirectional reconstruction loss \cite{lee2018diverse,huang2018multimodal} (requires carefully designed loss functions and selected hyper-parameters), thus making the training easier and faster. We expect more realistic results if adversarial training is involved, as well as more diverse synthesis if the output and the latent code is invertible.

\section{Experiments}

\subsection{Experimental Settings}
\noindent\textbf{Dataset.}
We conduct our experiments on DeepFashion (In-shop Clothes Retrieval Benchmark) dataset \cite{liu2016deepfashion} originally consisting of 52,712 person images with fashion clothes. In contrast to previous pose-guided generation approaches that use image pairs that contain people in the same clothes with two different poses for training and testing, we do not need paired data but rather images with multiple fashion items in order to model the compatibility among them. As a result, we filter the data and select 13,821 images that contains more than 3 fashion items to conduct our experiments. We randomly select 12,615 images as our training data and the other 1,206 for testing, while ensuring that there is no overlap in fashion items between two splits.

\noindent\textbf{Network Structure.}
Our shape generator and appearance generator share similar network structure. $G_s$ and $G_a$ have an input size of $256 \times 256$, and are built upon a U-Net \cite{ronneberger2015u} structure with 2 residual blocks \cite{he2015deep} in each encoding and decoding layer. We use convolutions with a stride of 2 to downsample the feature maps in encoding layers, and utilize nearest neighborhood interpolation to upscale the feature map resolution in the decoding layers. Symmetric skip connections \cite{ronneberger2015u} are added between encoder and decoder to enforce spatial correspondence between input and output. Based on the observations in \cite{zhu2017toward}, we set the length of all latent vectors to 8, and concatenate the latent vector to each intermediate layer in the U-Net after spatially replicating it to have the same spatial resolution. $E_s$, $E_{cs}$, $E_a$ and $E_{ca}$ all have similar structure as the U-Net encoder; except that their input is $128 \times 128$ and a fully-connected layer is employed at the end to output $\mu$ and $\sigma$ for sampling the Gaussian latent vectors. All convolutional layers have $3 \times 3$ kernels, and the number of channels in each layer is identical to \cite{pix2pix2016}. The detailed network structure is visualized in the supplementary material.

\noindent\textbf{Training Setup.}
Similar to recent encoder-decoder based generative networks, we use the Adam \cite{kingma2014adam} optimizer with $\beta_1=0.5$ and $\beta_2=0.999$ and a fixed learning rate of 0.0001. We train the compatible shape generator for 20K steps and the appearance generation network for 60K steps, both with a batch size of 16. We set $\lambda_{KL} = 0.1$ for both shape and appearance generators.

\begin{figure*}
\begin{center}
   \includegraphics[width = 1.0\linewidth]{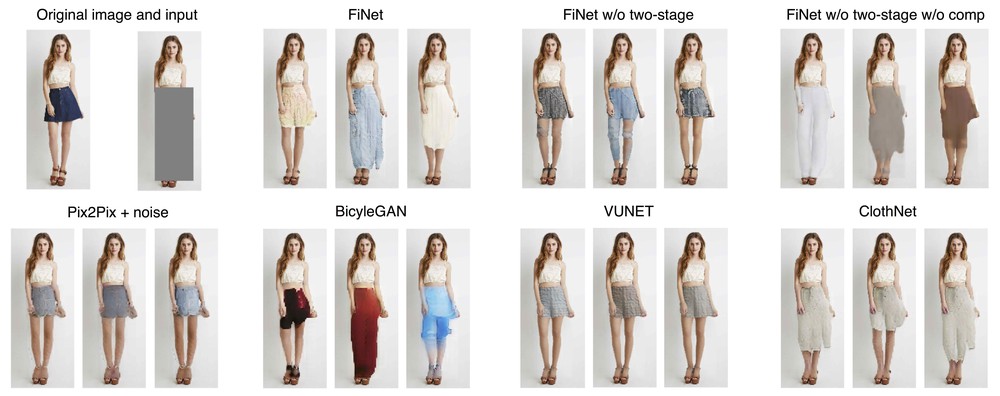}
\end{center}
\vspace{-20pt}
\caption{Inpainting comparison of different methods conditioned on the same input.}
\label{fig:comp}
\end{figure*}

\subsection{Compared Methods}
To validate the effectiveness of our method, we compare FiNet with the following methods:

\noindent\textbf{FiNet w/o two-stage}. We use a one-step generator to directly reconstruct image $I$ without the proposed two-stage framework. The one-step generator has the same network structure and loss function as our compatible appearance generator; the only difference is that it takes the pose heatmap, face and hair segment, $\hat S$ and $\hat I$ as input (\ie, merging the input of two stages into one).

\noindent\textbf{FiNet w/o comp}. Our method without compatibility encoder, \ie, minimizing $L_{KL}$ instead of $\hat L_{KL}$ in both shape and appearance generation networks.

\noindent\textbf{FiNet w/o two-stage w/o comp}. Our full method without two-stage training and compatibility encoder, which reduces FiNet to a one-stage conditional VAE \cite{kingma2013auto}.

\noindent\textbf{pix2pix + noise} \cite{pix2pix2016}. The original image-to-image translation frameworks are not designed for synthesizing missing clothing, thus we modify the input of this framework to have the same input as FiNet w/o two-stage. We add a noise vector for producing diverse results as in \cite{zhu2017toward}. Due to the inpainting nature of our problem, it can also be considered as a variant of a conditional context encoder \cite{pathak2016context}.

\noindent\textbf{BicyleGAN} \cite{zhu2017toward}. Because pix2pix can only generate single output, we also compare with BicyleGAN, which can be trained on paired data and output multimodal results. 
Note that we do not take multimodal unpaired image-to-image translation methods \cite{huang2018multimodal,lee2018diverse} into consideration since they usually produce worse results.

\noindent\textbf{VUNET} \cite{esser2018variational}. A variational U-Net that models the interplay of shape and appearance. We make the similar modification to the network input such that it models shape based on the same input as FiNet w/o two-stage and models the appearance using the target clothing appearance $x_a$.

\noindent\textbf{ClothNet} \cite{Lassner_2017_ICCV}. We replace the SMPL \cite{bogo2016keep} condition in ClothNet by our pose heatmap and reconstruct the original image. Note that ClothNet can generate diverse segmentation maps, but only outputs a single reconstructed image per segmentation map.

\noindent\textbf{Compatibility Loss}. Since most of the compared methods do not model the compatibility among fashion items, we also inject $x_c$ into these frameworks and design a compatibility loss to ensure that the generated clothing matches its contextual garments. It is similar to the loss of matching-aware discriminators in text-to-image synthesis \cite{reed2016generative,zhang2016stackgan} and can be easily injected into a generative model framework and trained end-to-end. Adding this loss to a network aims to inject compatibility information for fair comparison.

\begin{table*}[]
\ra{0.9}
\centering
\begin{tabular}{@{}lc*{5}c@{}}
\toprule
Method                            & Compatibility & Diversity & Human & IS \\ 
\midrule
Random Real Data &  1.000 & 0.676 $\pm$  0.086 &  50.0\%  & 3.907 $\pm$ 0.051 \\ 
\midrule

pix2pix  \cite{pix2pix2016}    &  0.628  &  0.057 $\pm$ 0.037 & 13.3\% & 3.629 $\pm$ 0.046   \\

BicyleGAN  \cite{zhu2017toward}          &  0.548    &   0.419 $\pm$  0.153&  16.4\%    & 3.596 $\pm$ 0.038    \\
VUNET  \cite{esser2018variational}          &  0.652    &   0.128 $\pm$ 0.096 &  30.7\% & 3.559 $\pm$ 0.041         \\
ClothNet \cite{Lassner_2017_ICCV}      &  0.621     &  0.212 $\pm$ 0.127 &     15.9\%     & 3.573 $\pm$ 0.058   \\ 
\midrule

FiNet w/o two-stage w/o comp & 0.513 & 0.417 $\pm$ 0.128 &   12.8\%    & 3.681 $\pm$ 0.050  \\
FiNet w/o comp &  0.528 &  \textbf{0.424 $\pm$ 0.125} &  12.3\%   &  \textbf{3.688 $\pm$ 0.030}   \\
FiNet w/o two-stage &  0.666 &  0.261 $\pm$ 0.144   &   25.6\%  & 3.570 $\pm$ 0.046    \\
FiNet (Ours full) &  \textbf{0.683}   &  0.297 $\pm$ 0.141 &   \textbf{36.6\%}     & 3.564 $\pm$ 0.043 \\

\bottomrule
\end{tabular}
\vspace{-5pt}
\caption{Quantitative comparisons in terms of compatibility, diversity and realism.}
\label{tab:comp}
\end{table*}

\subsection{Qualitative Results}
In Figure \ref{fig:comp}, we show 3 generated images of each method conditioned on the same input. We can see that FiNet generates visually compatible bottoms with different shapes and appearances. Without generating the semantic layout as intermediate guidance, FiNet w/o two-stage cannot properly determine the clothing boundaries. FiNet w/o two-stage w/o comp also produces boundary artifacts and the generated appearances do not match the contextual garments. pix2pix \cite{pix2pix2016} + noise only generates results with limited diversity---it tends to learn the average shape and appearance based on distributions of the training data. BicyleGAN \cite{zhu2017toward} improves diversity, but the synthesized images are incompatible and suffer from artifacts brought by adversarial training. We found VUNET suffers from posterior collapse and only generates similar shapes. ClothNet \cite{Lassner_2017_ICCV} can generate diverse shapes but with similar appearances because it also uses a pix2pix-like structure for appearance generation.

We show more results of our proposed FiNet in Figure \ref{fig:teaser}, which further illustrates the effectiveness of FiNet for generating different types of garments with high compatibility and diversity. Note that FiNet is also able to generate fashion items that do not exist in the original image as shown in the last example in Figure \ref{fig:teaser}.

\begin{figure*}
\begin{center}

   \includegraphics[width = 1.0\linewidth]{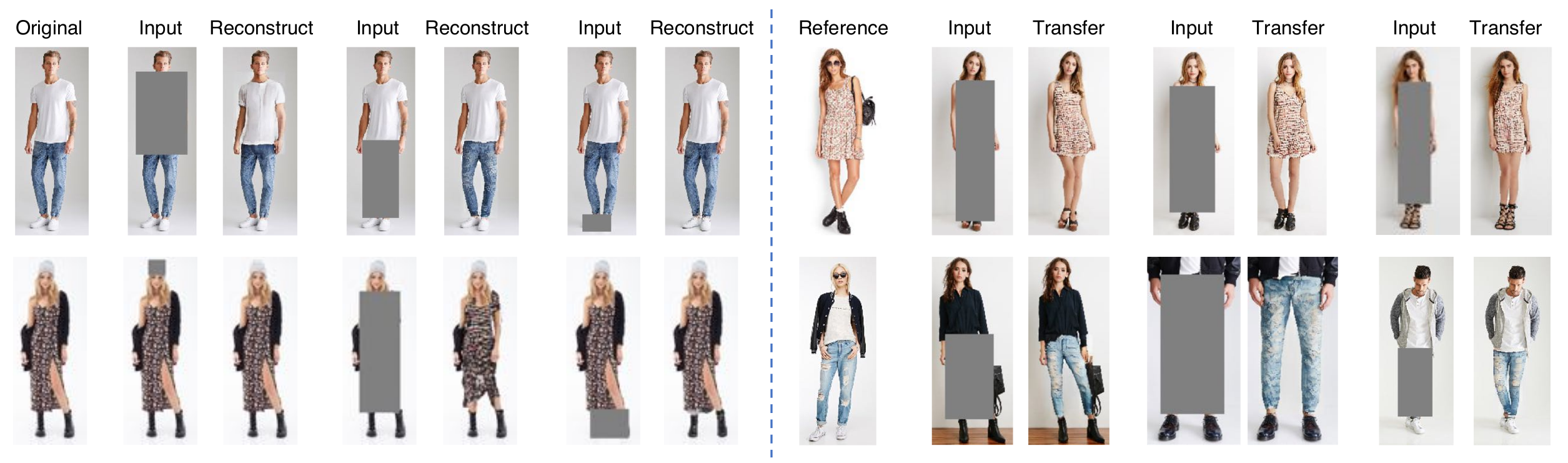}
\end{center}
\vspace{-16pt}
\caption{Conditioning on different inputs, FiNet can achieve clothing reconstruction (left), and clothing transfer (right).}
\label{fig:rec}
\end{figure*}

\subsection{Quantitative Comparisons}
We compare different methods in terms of compatibility, diversity and realism.

\noindent \textbf{Compatibility}. 
To properly evaluate the compatibility of generated images, we trained a compatibility predictor adopted from \cite{veit2015learning}. The training labels also come from the same weakly-supervised compatibility assumption---if two fashion items co-occur in a same catalog image, we regard them as a positive pair, otherwise, these two are considered as negative \cite{song2018neural,han2017learning}. We fine-tune an Inception-V3 \cite{Szegedy2015} pre-trained on ImageNet on DeepFashion training data for 100K steps, with an embedding dimension of 512 and default hyper-parameters. We use the RGB clothing segments as input to the network. Following \cite{esser2018variational,raj2018swapnet} that measure visual similarity using feature distance in a pretrained VGG-16 network, we measure the compatibility between a generated clothing segment and the ground truth clothing segment by their cosine similarity in the learned 512-D compatibility embedding space. 

\noindent\textbf{Diversity}. Besides compatibility, diversity is also a key performance metric for our task. Thus, we utilize LPIPS \cite{zhang2018perceptual} to measure the diversity of generated images (only inpainted regions) as in \cite{zhu2017toward,lee2018diverse,huang2018multimodal}. 2,000 image pairs generated from 100 fashion images are used to calculate LPIPS.

\noindent\textbf{Realism}. 
We conduct a user study to evaluate the realism of generated images. Following
\cite{chen2017cascaded,ma2017pose,zhu2017toward}, we perform time-limited (0.5s) real or synthetic test with 10 human raters. The human fooling rate (chance is 50\%, higher is better) indicates realism of a method. As a popular choice, we also report Inception Score \cite{salimans2016improved} (IS) for evaluating realism.

We make the following observations based on the comparisons present in Table \ref{tab:comp}:

(1) FiNet yields the highest compatibility score with meaningful diversity. Without the compatibility-aware KL regulation, BicyleGAN and our baselines w/o comp fail to inpaint compatible clothes. These methods learn to project all potential garments into the same distribution independent of contextual garments, resulting in incompatible but highly diverse images. In contrast, VUNET, ClothNet give higher compatibility while sacrificing diversity. pix2pix ignores the injected noise and cannot generate diverse outputs. Their quantitative performances match their qualitative behaviors given in Figure \ref{fig:comp}.

(2) Our method achieves the highest human fooling rate. We find that Inception score does not correlate well with the human perceptual study, making it unsuitable for our task. This is because IS tends to reward image content with rich textures and large diversity. For example, colorful shoes usually look inharmonious but have a high IS. Similar observations have also been made in \cite{raj2018swapnet,han2018viton}. 

(3) Images with low compatibility scores usually have lower human fooling rates. This confirms that incompatible garments also looks unrealistic to human.

\subsection{Clothing Reconstruction and Transfer}
\label{sec:rec}

Trained with a reconstruction loss, FiNet can also be adopted as a two-stage clothing transfer framework. More specifically, for an arbitrary target garment $t$ with shape $t_s$ and appearance $t_a$, $G_s(\hat S, p_s, t_s)$ can generate the shape of $t$ in the missing region of $\hat S$, while $G_a(\hat I, p_a, t_a)$ can synthesize an image with the appearance of $t$ filling in $\hat I$. This can produce promising results for clothing reconstruction (when $t = x$, where $x$ is the original missing garment) and garment transfer (when $t\neq x$) as shown in Figure \ref{fig:rec}. FiNet inpaints the shape and appearance of target garment natually onto a person, which further demonstrates its ability in generating realistic fashion images.

\section{Conclusion}
We introduce FiNet, a two-stage generation network for synthesizing compatible and diverse fashion images. By decomposition of shape and appearance generation, FiNet can inpaint garments in target region with diverse shapes and appearances. Moreover, we integrate a compatibility module that encodes compatibility information to the network, constraining the generated shapes and appearances to be close to the existing clothing pieces in a learned latent style space. The superior performance of FiNet suggests that it can be potentially used for compatibility-aware fashion design and new fashion item recommendation.

\section*{Acknowledgement}
Larry S. Davis and Zuxuan Wu are partially supported by the Office of Naval Research under Grant N000141612713.

{\small
\bibliographystyle{ieee}
\bibliography{egbib}
}

\clearpage

\appendix

\section{Network Details}
\begin{figure*}[h]
\begin{center}
   \includegraphics[width = 1.0\linewidth]{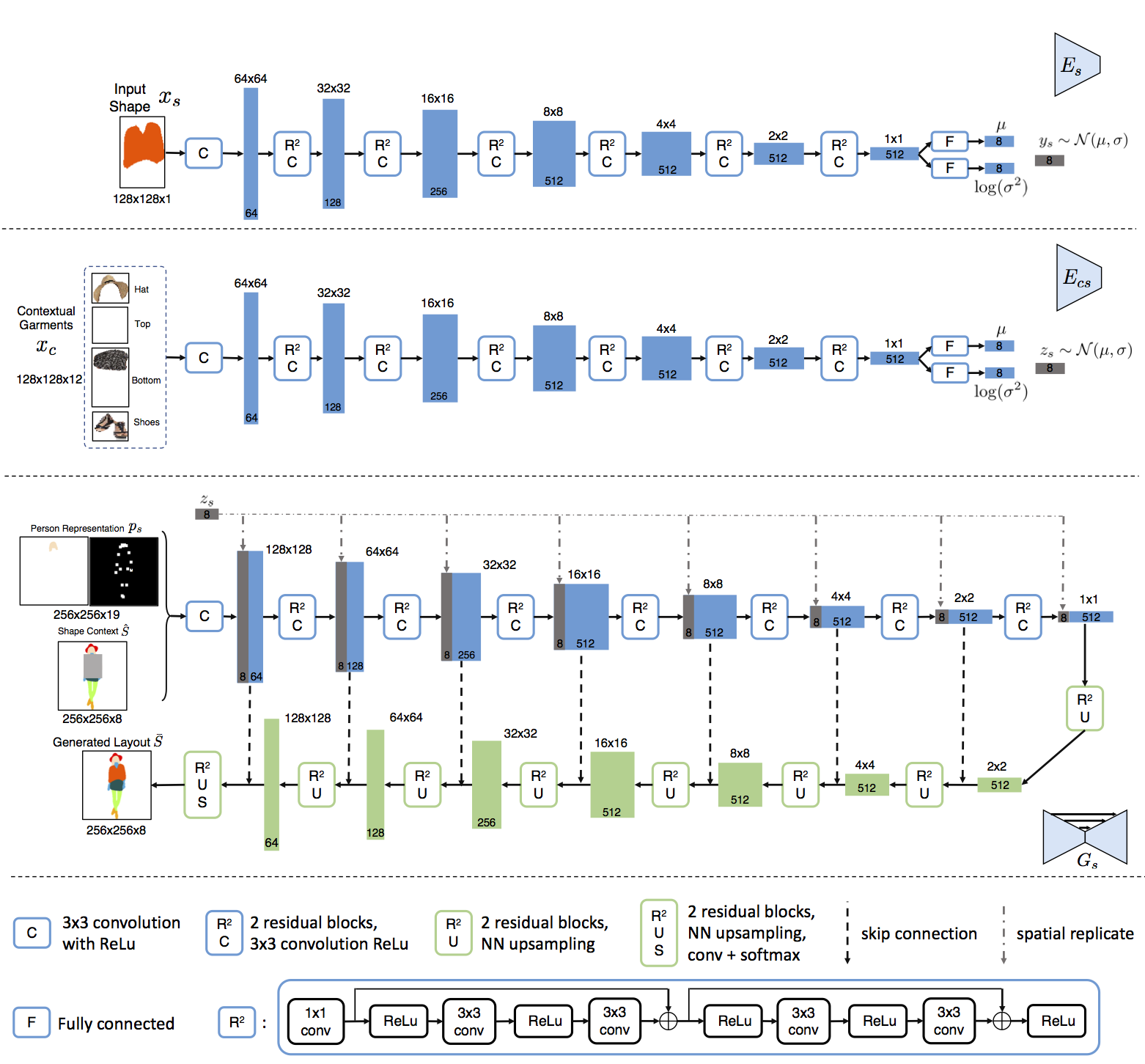}
\end{center}
\vspace{-10pt}
\caption{Network structure of our shape generation network.}
\label{fig:s1}
\end{figure*}

\begin{figure*}[h]
\begin{center}
   \includegraphics[width = 1.0\linewidth]{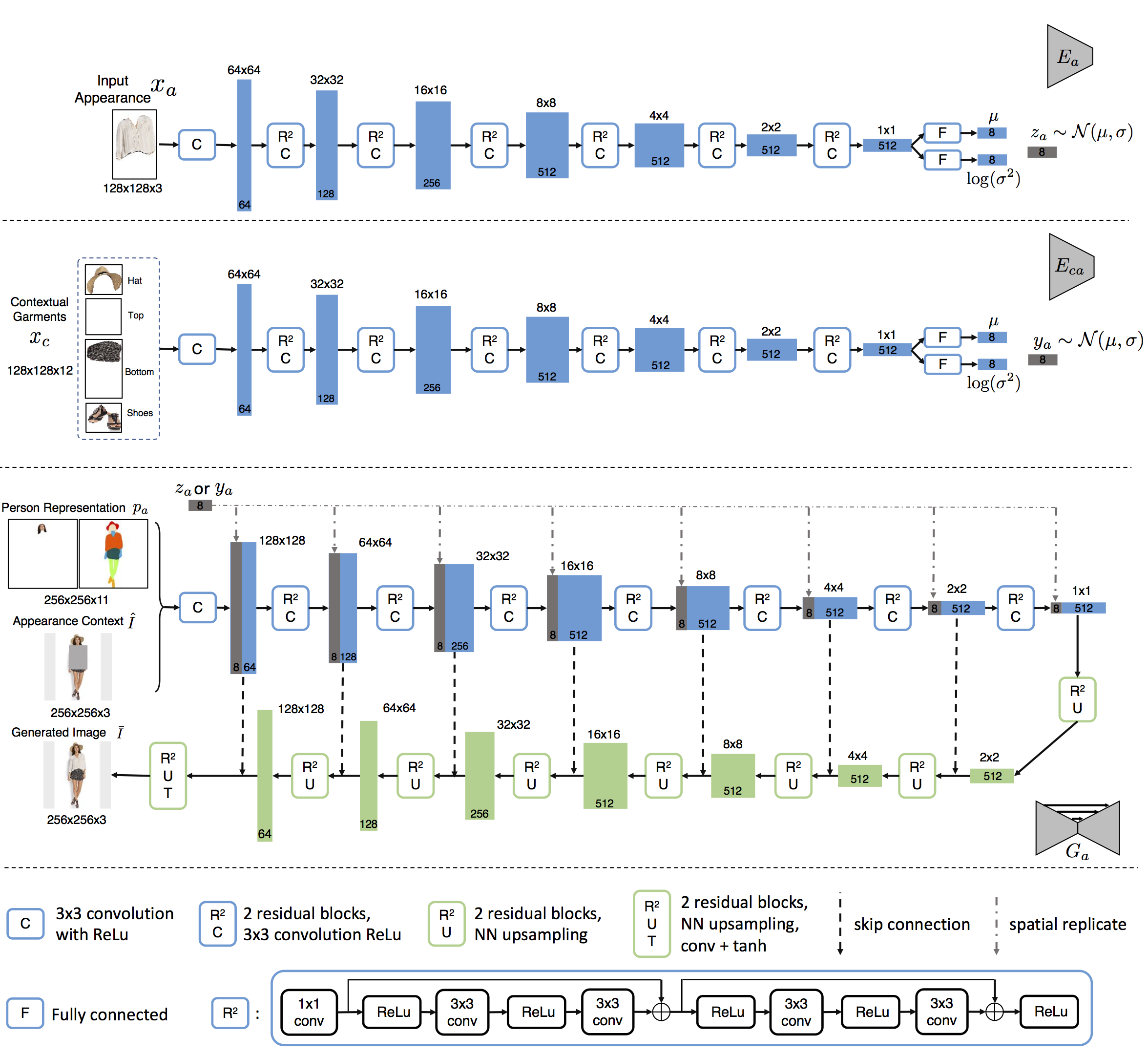}
\end{center}
\vspace{-10pt}
\caption{Network structure of our appearance generation network.}
\label{fig:s2}
\end{figure*}

We illustrate the detailed network structures of our shape generation network and appearance generation network in Figure \ref{fig:s1} and Figure \ref{fig:s2}, respectively. There are some details to be noted:

$\bullet$ Our residual block module $R^2$ contains two residual blocks \cite{he2015deep} with one $1\times1$ convolution in the beginning to make the number of input and output channels consistent after concatenation with the latent vector. We use $3\times3$ convolution for all the other convolutional operations in our network.

$\bullet$ We use the softmax function at the end of our shape generation network to generate segmentation maps, and the \textit{tanh} activation is applied when our appearance generation network outputs synthesized images. Other activations utilize \textit{ReLu}. No batch normalization is used in our network.

$\bullet$ To create layouts / images with a missing fashion item (\ie, shape context $\hat S$ and appearance context $\hat I$), we need to mask out pixels of a fashion item, which is determined by the plausible region of a specific garment category. Given the human parsing results generated by \cite{gong2018instance}, for a top item, we mask out the bounding box covering the regions of the top and upper body skin; for a bottom item, its plausible region contains bottom and lower body skin; for both hats and shoes, we use the bounding box of the corresponding fashion item to decide which region to mask out. The bounding boxes are slightly enlarged to ensure full coverage of these regions.

$\bullet$ For both shape and appearance, the inpainted region is first resized to $256\times256$, and the reconstruction losses are only computed over this resized inpainted region. Finally, we paste this region back to the input image and obtain the final result.

$\bullet$ For constructing contextual garments, we first utilize human parsing results $S$, generated by \cite{gong2018instance}, to extract an image segment for each garment in its corresponding plausible region. Then, we resize these extracted image segments to $128\times128$, and concatenate them in the order of \textit{hat}, \textit{top}, \textit{bottom}, \textit{shoes} with the target garment category one set to all $1$'s (\eg, $top$ is the target category in Figure \ref{fig:s1} and \ref{fig:s2}). This not only encodes the information of all contextual garments but also tells the network which category is missing. Finally, we have a $128\times 128 \times 12$ contextual garment representation $x_c$.

\section{Latent Space Visualization}

\begin{figure*}[h]
\begin{center}
   \includegraphics[width = 1.0\linewidth]{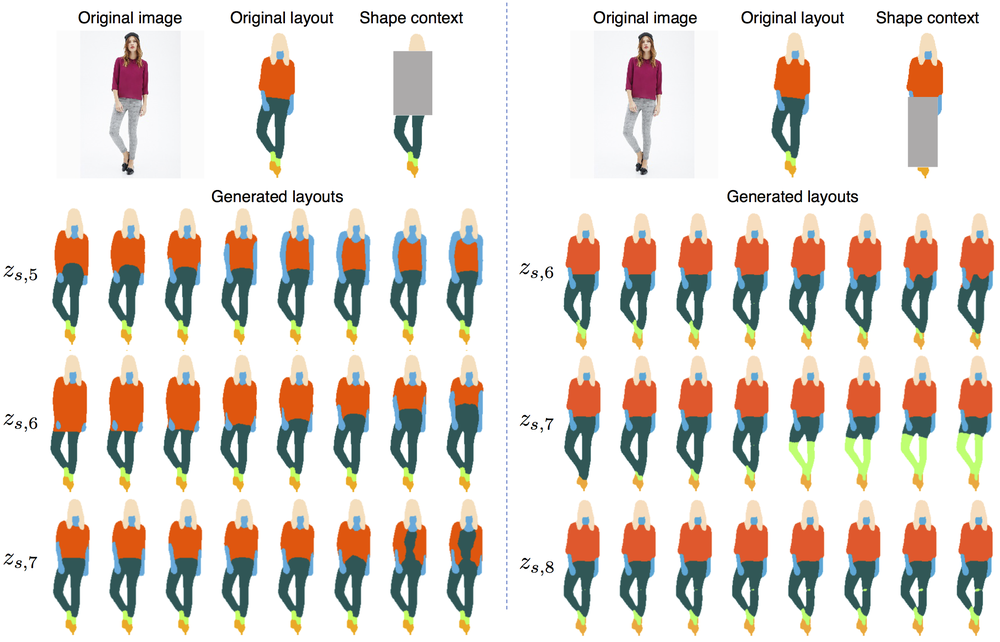}
\end{center}
\vspace{-10pt}
\caption{Generated layouts by our shape generation network when we change the values in different dimensions of the learned latent vector.}
\label{fig:shape_latent}
\end{figure*}

\textbf{Shape latent space}. In Figure \ref{fig:shape_latent}, we visualize the generated segmentation maps of an image when varying different dimensions of the shape latent vector $z_s$. In the left example, the shape generation network generates different top garment layouts when we change the values of the $5$-th, $6$-th, and $7$-th dimension of $z_s$ (note that they are also the corresponding dimensions of $y_s$ since $y_s$ and $z_s$ share the same latent space). We can find that different dimension controls different characteristics of the generated layouts: the $5$-th dimension mainly controls the sleeve length---long sleeve $\rightarrow$ middle sleeve  $\rightarrow$  short sleeve  $\rightarrow$ sleeveless; the $6$-th dimension determines the length of the clothing as well as the sleeve; and the $7$-th dimension measures if the top opens in the middle. As for the right example, in which we generate bottom garment, the $6$-th dimension is related to how the bottom interacts with the top; the length of the pants are changed when we vary the $7$-th dimension; and the last dimension correlates with the exposure of the knee. Note that, for different garment categories, the same dimension of $z_s$ (or $y_s$) controls different characteristics. For example, $z_{s,7}$ has something to do with the length of bottoms but not the length of tops.

\begin{figure*}[!t]
\begin{center}
   \includegraphics[width = 1.0\linewidth]{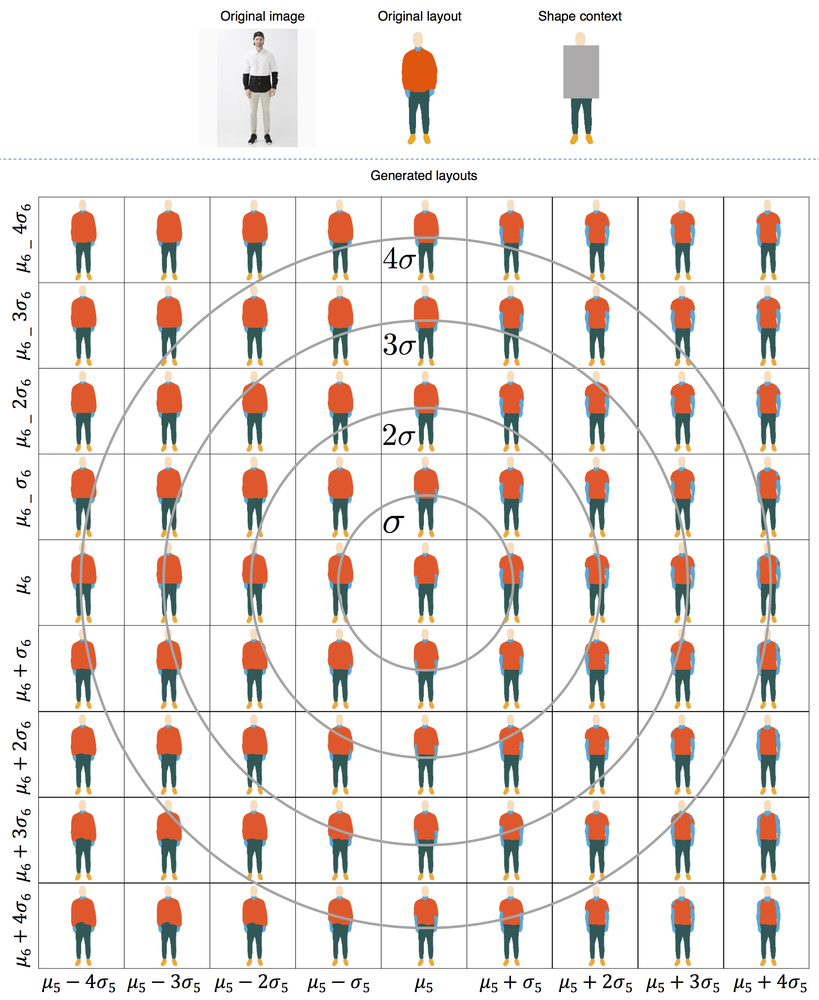}
\end{center}
\vspace{-10pt}
\caption{Shape compatibility space visualization. $x$ and $y$ axes correspond to the $5$-th and $6$-th dimensions of $y_s$, respectively.}
\label{fig:shape_gaussian1}
\end{figure*}

\begin{figure*}[!t]
\begin{center}
   \includegraphics[width = 1.0\linewidth]{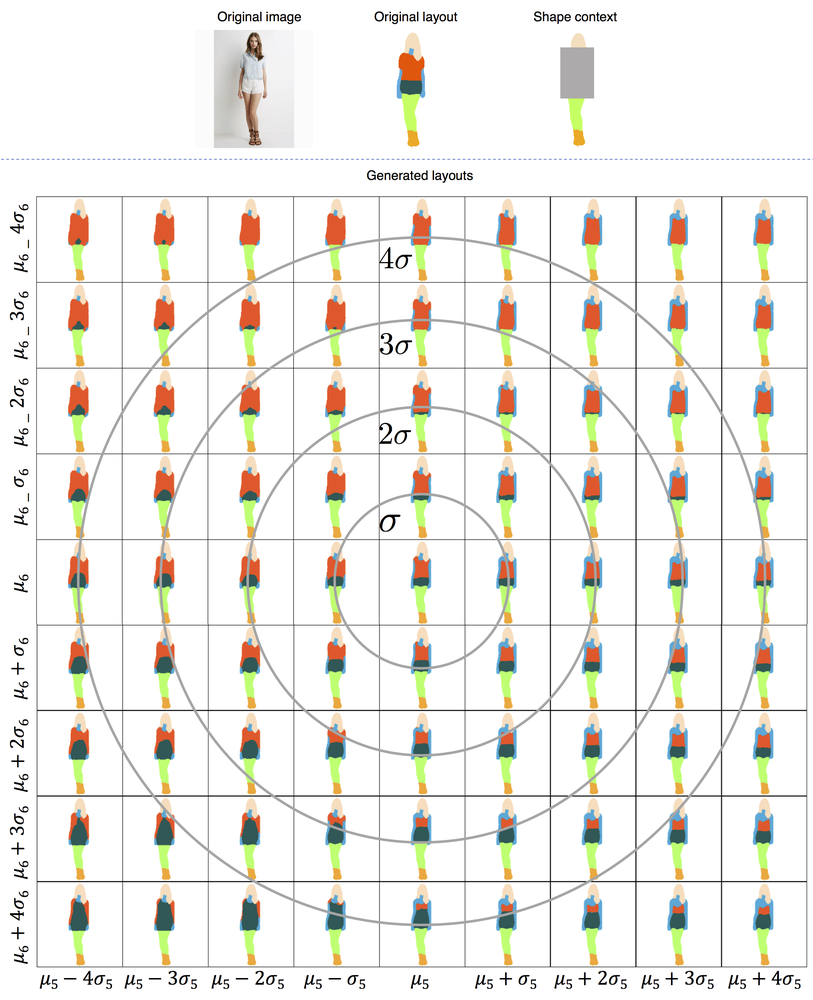}
\end{center}
\vspace{-10pt}
\caption{Shape compatibility space visualization. $x$ and $y$ axes correspond to the $5$-th and $6$-th dimensions of $y_s$, respectively.}
\label{fig:shape_gaussian2}
\end{figure*}

Further, in Figure \ref{fig:shape_gaussian1} and \ref{fig:shape_gaussian2}, we show the compatibility space for two images by projecting the generated layouts in a 2D plane, whose $x$ and $y$ axes correspond to the $5$-th (sleeve length) and $6$-th (clothing length) dimension of $y_s$ that is used to generate these layouts. $\mu_i$ and $\sigma_i$ represent the mean and standard deviation of $y_s$'s distribution in its $i$-th dimension. Consequently, layouts corresponding to latent vectors that are far from $\mu$ are uncompatible and unlikely to be generated.

In Figure \ref{fig:shape_gaussian1}, as we want to generate a compatible top for a man wearing a pair of long pants, the generated top layouts usually have long or short sleeves; and sleeveless tops (images in the lower right corner) are less compatible and realistic, thus these layouts are not likely to be generated (outside of $3\sigma$). In contrast, when we generate layouts for a girl with a pair of shorts as shown in Figure \ref{fig:shape_gaussian2}, the generated layouts tend to have shorter sleeve length as well as clothing length because they are more compatible with shorts. By the comparison between Figure \ref{fig:shape_gaussian1} and \ref{fig:shape_gaussian2}, we can see that our shape generation network effectively models the compatibility and can generate compatible garment shapes according to contextual information.

There are some unrealistic sleeve shapes presented in Figure \ref{fig:shape_gaussian1}. Due to the continuous nature of Normal distribution, one cannot guarantee that every code in the latent space corresponds to a realistic shape. We tried to add adversarial training to the shape generation network, but found it is still hard to rule out all unrealistic cases. A potential solution is to have a discrete latent space with finite samples, which, however, will need more complicated modeling (\eg, \cite{hu2018learning}). This is beyond the scope of this paper, and we consider it as a future research direction. Note that many of these unrealistic shapes fall out of $3\sigma$. This indicates that unrealistic shapes are often incompatible and can be avoided to some extent by our shape generation network.

\begin{figure*}[h]
\begin{center}
   \includegraphics[width = 1.0\linewidth]{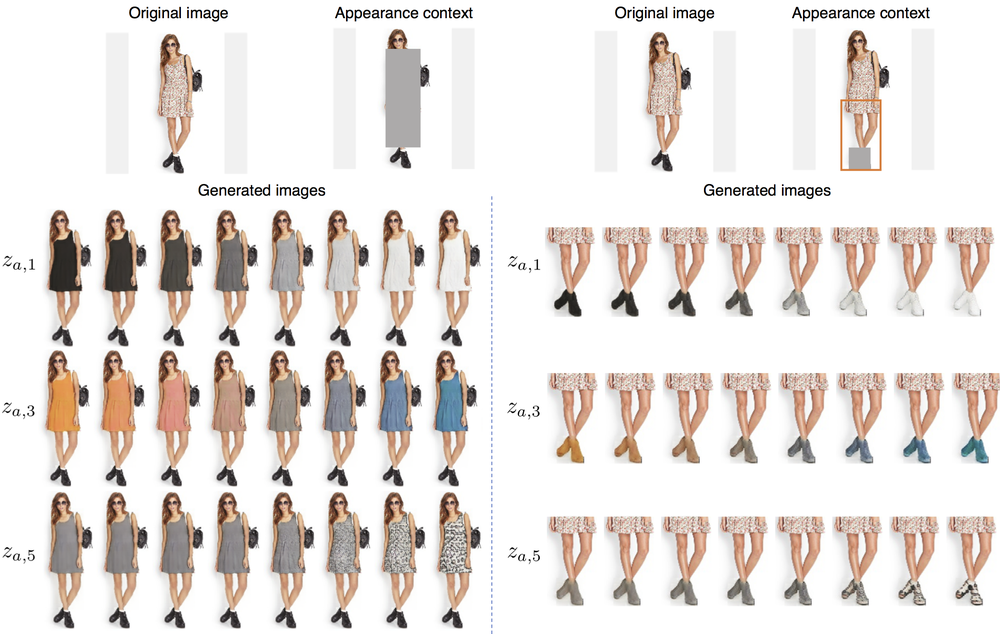}
\end{center}
\vspace{-10pt}
\caption{Generated layouts by our appearance generation network when we change the values in different dimensions of the learned latent vector.}
\label{fig:appearance_latent}
\end{figure*}

\textbf{Appearance latent space}. As shown in Figure \ref{fig:appearance_latent}, we further present similar visualization for the appearance latent vector $z_a$ (or $y_a$). Note that for visualization purposes, we use the ground truth segmentation map to generate appearance for simplicity. Unlike shape, the same dimension of the appearance latent vector correlates to the same appearance characteristic for different garment categories. The $1$-st, $3$-rd, $5$-th dimensions correspond to brightness, color, texture of the generated images, respectively. This also indicates the importance of learning a compatible space; otherwise, if we project all appearances in the same latent space as FiNet w/o comp or BicyleGAN \cite{zhu2017toward} without conditioning on the contextual garments, incompatible and visually unappealing shoes (shoes of all different colors as in the right side of Figure \ref{fig:appearance_latent}) may be generated.

\begin{figure*}[!t]
\begin{center}
   \includegraphics[width = 0.98\linewidth]{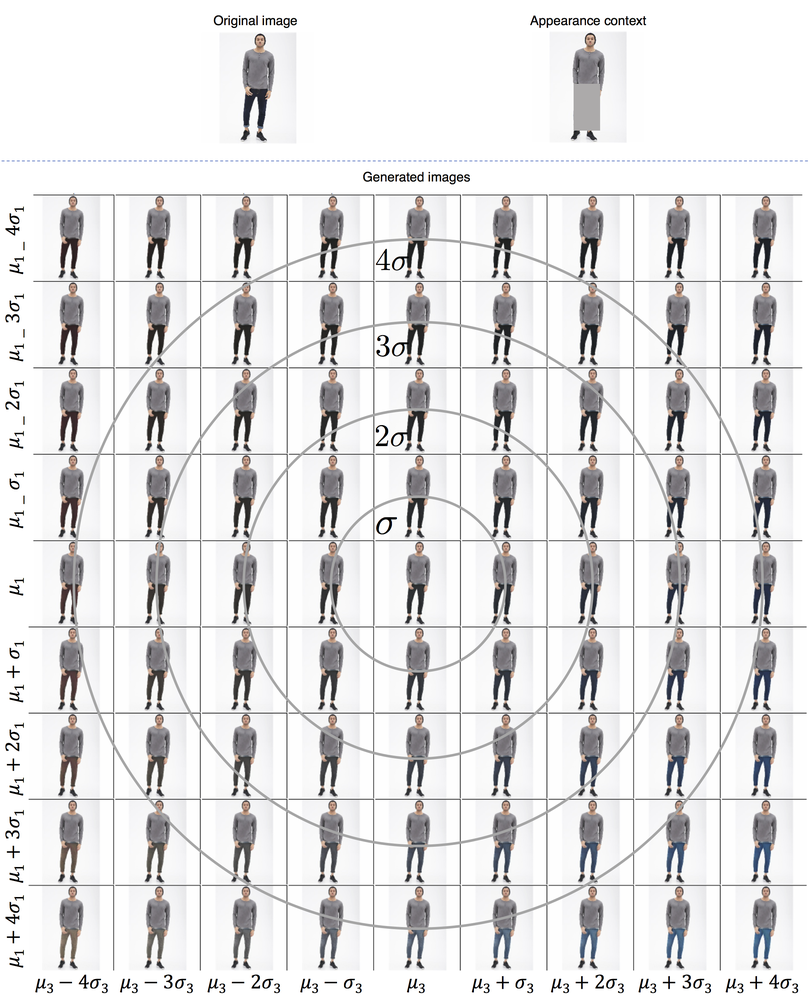}
\end{center}
\vspace{-10pt}
\caption{Appearance compatibility space visualization. $x$ and $y$ axes correspond to the $3$-rd and $1$-st dimensions of $y_a$, respectively.}
\label{fig:appearance_gaussian1}
\end{figure*}

\begin{figure*}[!t]
\begin{center}
   \includegraphics[width = 0.98\linewidth]{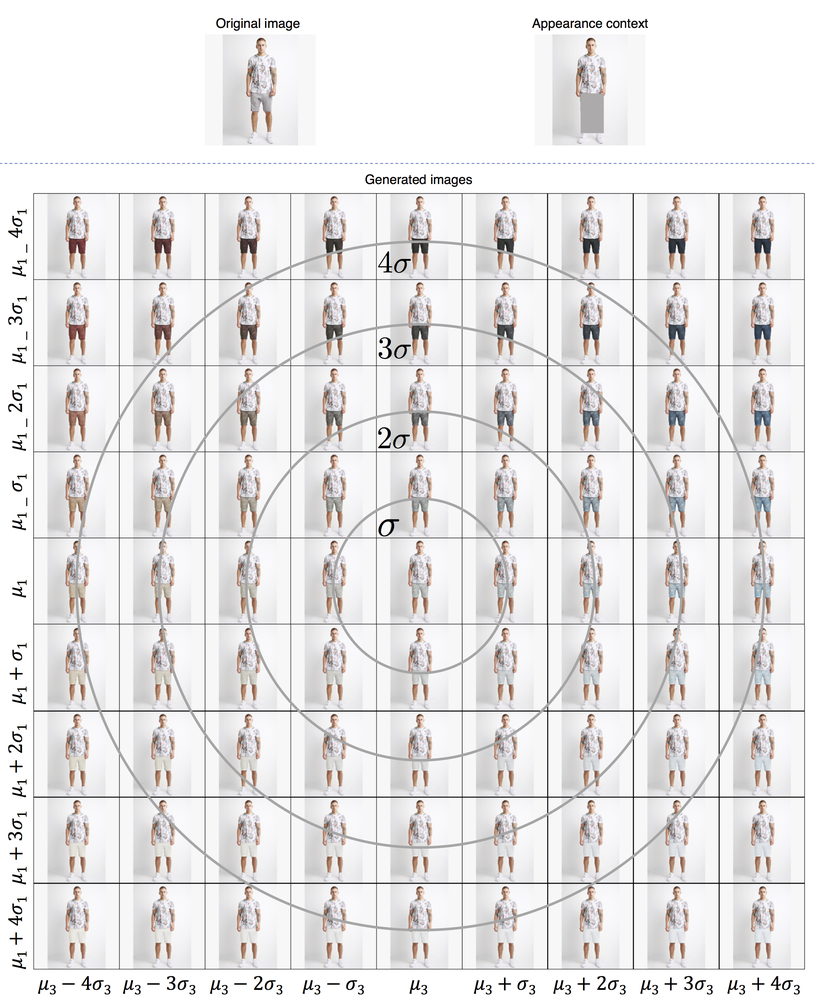}
\end{center}
\vspace{-10pt}
\caption{Appearance compatibility space visualization. $x$ and $y$ axes correspond to the $3$-rd and $1$-st dimensions of $y_a$, respectively.}
\label{fig:appearance_gaussian2}
\end{figure*}

\begin{figure*}[!t]
\begin{center}
   \includegraphics[width = 0.98\linewidth]{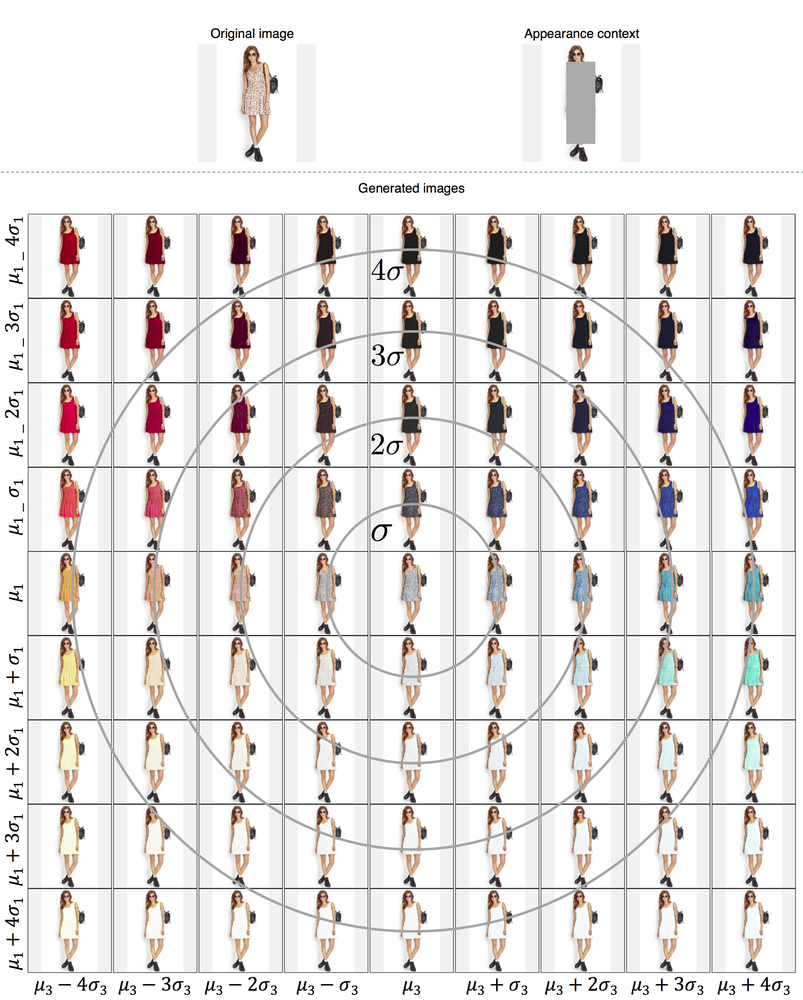}
\end{center}
\vspace{-10pt}
\caption{Appearance compatibility space visualization. $x$ and $y$ axes correspond to the $3$-rd and $1$-st dimensions of $y_a$, respectively.}
\label{fig:appearance_gaussian3}
\end{figure*}

We further plot the appearance compatibility space for three exemplar images in Figure \ref{fig:appearance_gaussian1}, \ref{fig:appearance_gaussian2} and \ref{fig:appearance_gaussian3} for better understanding our appearance generation network. The generated appearances are arranged according to the $1$-st (brightness) and $3$-rd (color) dimension of $y_a$. In particular, in Figure \ref{fig:appearance_gaussian1}, given the gray top, our network considers dark bottoms of black or blue as compatible, and the ground truth pants also present these visual characteristics. In Figure \ref{fig:appearance_gaussian2}, we can find that since the white graphic T-shirt is more compatible with lighter bottoms, our generation network creates such shorts accordingly. Unlike these two cases, in Figure \ref{fig:appearance_gaussian3}, there is no strong constraint in the color of a compatible dress, so the appearance generation network outputs dresses with various colors. The results illustrated in these figures again validate that we inject compatibility information into our network to ensure diverse and compatible image inpainting results.

\section{Multimodal Advantages of FiNet}

\begin{table}[h]
\centering
\small
\begin{tabular}{@{}lc*{4}c@{}}

\toprule
Method                            & Comp & Diversity  & IS \\ 
\midrule

BicyleGAN \cite{zhu2017toward}         &  0.568    &   0.29 $\pm$ 0.13    & 3.62 $\pm$ 0.04    \\
VUNET  \cite{esser2018variational}          &  0.637    &   0.30  $\pm$  0.11 & 3.59  $\pm$  0.04         \\
ClothNet \cite{Lassner_2017_ICCV}      &  0.644     &  0.29  $\pm$  0.12     & 3.60  $\pm$  0.06   \\ 
\midrule

FiNet w/o 2-stage w/o comp & 0.548 & 0.29 $\pm$ 0.11     & 3.52 $\pm$ 0.05  \\
FiNet w/o comp &  0.583 &  0.31 $\pm$ 0.12   &  3.57 $\pm$ 0.04   \\
FiNet w/o two-stage &  0.656 &  0.29 $\pm$ 0.13  & 3.51 $\pm$ 0.05    \\
FiNet (Ours full) &  \textbf{0.683}   &  0.30 $\pm$ 0.14    & 3.56 $\pm$ 0.04 \\

\bottomrule
\end{tabular}
\caption{Comparisons of different methods with similar diversity.}
\label{tab:supp_comp}
\vspace{-30pt}
\end{table}

In Table 1 of the main paper, we showed the quantitative comparisons in terms of compatibility, diversity and realism of different methods. However, there should be some trade-off between diversity and realism, which could be controlled by adjusting the sampling of latent vectors at test time. To this end, to better demonstrate the multimodal advantage of our method, we conduct experiments by adjusting all methods to have a similar diversity (LPIPS metric). This is done by changing the input latent code (\eg, increasing or decreasing $\sigma$ of the latent code during inference) and fed it into the generator. We present this experimental results in Table \ref{tab:supp_comp}. We do not include new results for pix2pix+noise, because merely adding a noise is hard to make pix2pix to have such a large LPIPS value while maintaining realistic generation. Moreover, as shown in the Table 1 in the main paper, FiNet already achieves higher compatibility, diversity and human fooling rate than pix2pix+noise. Comparing Table \ref{tab:supp_comp} with the original results shown in the main paper, we find that the trade-off exists. For example, for FiNet w/o 2-stage w/o comp and FiNet w/o comp, when the diversity decreases, they get a bit higher compatibility and realism. Plus, FiNet still achieves higher compatibility and comparable IS with similar diversity.

Moreover, compared to other VAE-like methods, FiNet does not assume a single fixed distribution $\mathcal{N}(0,\mathbbm{1})$, but conditions the distribution on the compatible contextual garments (different distributions given different $x_c$), which not only effectively injects the compatibility but also prevents collapsing to a single distribution. During experiments, we find that FiNet has a higher and more stable KLD and mutual information (MI) between latent code and observation than a vanilla VAE; larger values of KLD and MI suggest the active use of latent variables and robustness to posterior collapse \cite{he2019lagging}.

\section{Clothing Reconstruction and Transfer}

\begin{figure*}[h]
\begin{center}
   \includegraphics[width = 1.0\linewidth]{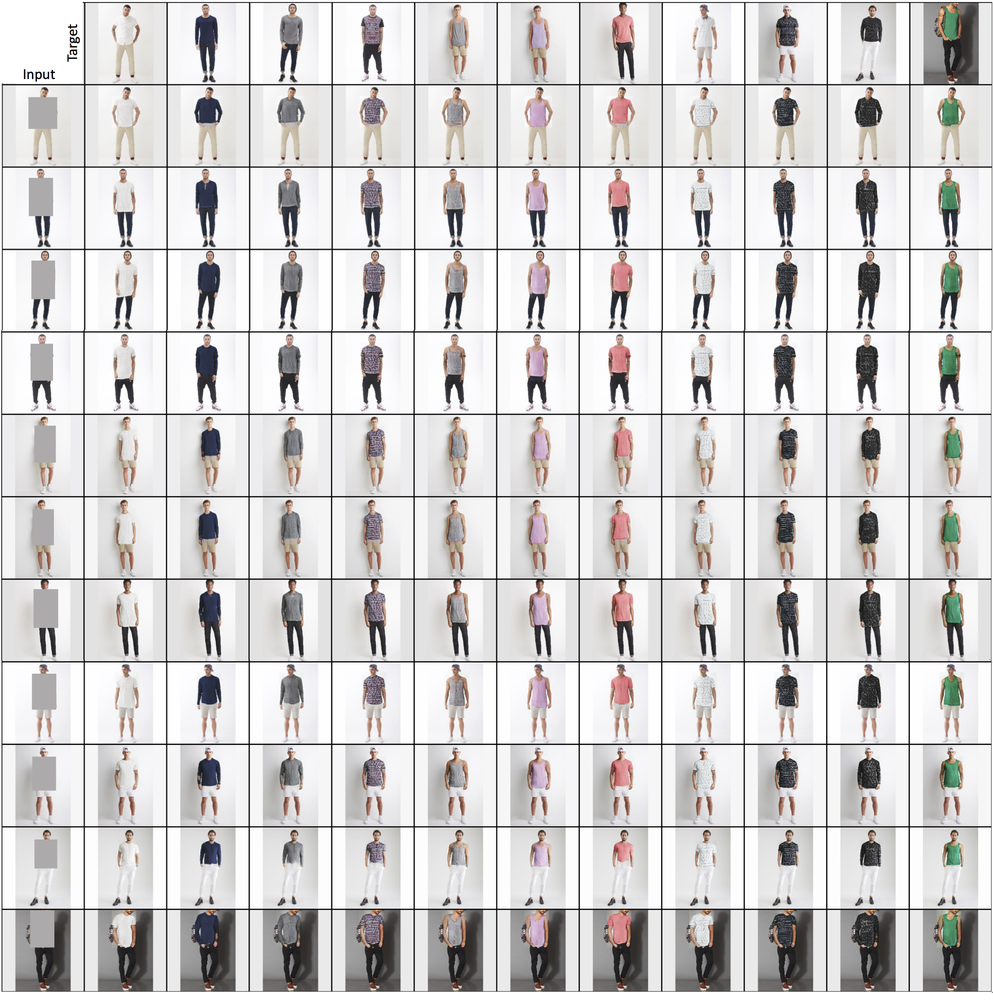}
\end{center}
\vspace{-10pt}
\caption{Clothing transfer results of tops. Each row corresponds to an input image whose top garment is transfered from different target tops. The diagonal images are reconstruction results, since the input and target images are the same. FiNet can naturally render the shape and appearance of the target garment onto other people with various poses and body shapes.}
\label{fig:transfer1}
\end{figure*}

\begin{figure*}[h]
\begin{center}
   \includegraphics[width = 1.0\linewidth]{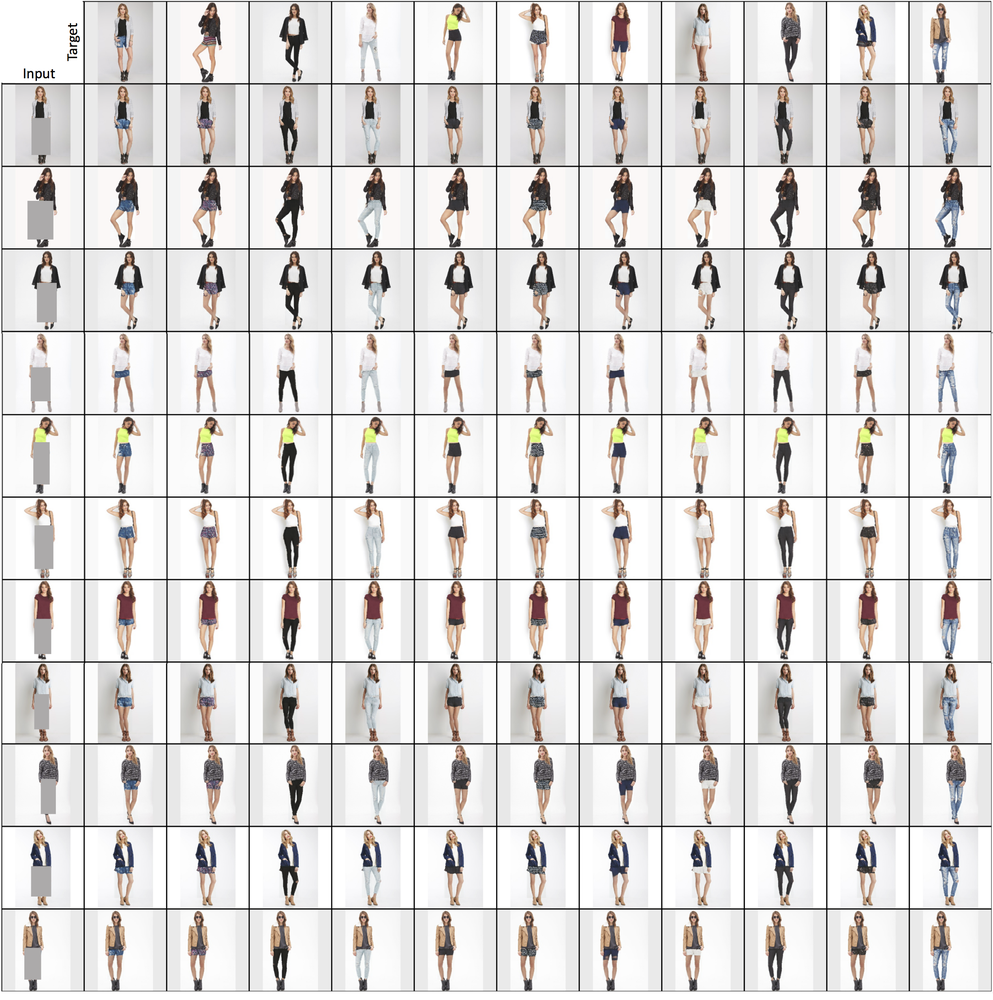}
\end{center}
\vspace{-10pt}
\caption{Clothing transfer results of bottoms. Each row corresponds to an input image whose bottom garment is transfered from different target bottoms. The diagonal images are reconstruction results, since the input and target images are the same. FiNet can naturally render the shape and appearance of the target garment onto other people with various poses and body shapes.}
\label{fig:transfer2}
\end{figure*}

Although clothing reconstruction and transfer is not our main contribution, we show more transfer results in Figure \ref{fig:transfer1} and \ref{fig:transfer2} to demonstrate that our method, by reconstructing a target garment and fill it in the missing regions of an input image, can transfer the target garment naturally to the input image. Note that FiNet transfers shape and appearance by inpainting a specific clothing item, which is different from most existing approaches that generate the full person as a whole \cite{ma2017pose,esser2018variational,han2018viton}. This potentially provides a new solution for applications like virtual try-on \cite{han2018viton} and generating people in diverse clothes \cite{Lassner_2017_ICCV}.

\end{document}